# Knowledge Retrieval

Naseem Shaik
Department of Computer Science and Engineering
University of South Florida
Tampa, Florida
naseemshaik@usf.edu

*Abstract*—**Robots can complete all human-performed tasks, but due to their current lack of knowledge, some tasks still cannot be completed by them with a high degree of success. However, with the right knowledge, these tasks can be completed by robots with a high degree of success, reducing the amount of human effort required to complete daily tasks. In this paper, the FOON, which describes the robot action success rate, is discussed. The functional object-oriented network (FOON) is a knowledge representation for symbolic task planning that takes the shape of a graph. It is to demonstrate the adaptability of FOON in developing a novel and adaptive method of solving a problem utilizing knowledge obtained from various sources, a graph retrieval methodology is shown to produce manipulation motion sequences from the FOON to accomplish a desired aim. The outcomes are illustrated using motion sequences created by the FOON to complete the desired objectives in a simulated environment.**

## I. INTRODUCTION

Robotic cooking performance varies depending on the recipe being executed. Depending on the inputs, ingredients, and objects as well as the order in which they are delivered, they keep altering. The robot can become confused as a result, making it unsure of how to carry out the task. Therefore, to avoid confusion, we use knowledge representation in terms of FOON, and by creating FOON-specific search algorithms, we make it apparent how the robot should go about carrying out the tasks.

Previous studies had a smaller data collection, therefore the robot could only carry out the tasks that were predetermined in the data set at the time. Robots only have a limited amount of information, which causes them to be perplexed by unexpected ingredient combinations. Say, for instance, that a robot has a formula for producing a smoothie with specific ingredients; however, when a new ingredient is introduced, the robot is unsure of what to do and how to proceed with the task because it does not have that knowledge in its knowledge representation. Therefore, when new data is provided, the robot must train itself using the previously collected data and take the necessary action to complete the task.

## II. VIDEO ANNOTATION AND FOON CREATION

Manual video annotation involves watching the recipe video and taking notes on the items, their states, and the motions that were utilized in each step. A functional unit defines how the states of the objects involved in a manipulation action change before and after execution, where states can be used to measure the success of an operation. Both output object nodes and input object nodes specify the state(s) of the objects that must be in order for the job to be completed, respectively. There might be fewer output object nodes than input object nodes in some situations because some operations might not affect the states of all input objects. A subgraph is a FOON that depicts a single activity. It has functional units in order to describe the states of the objects before and after each action, the timestamps at which each action occurs in the source demonstration, and the objects that are being modified. A subgraph that is obtained from knowledge retrieval is called a task tree.

For illustrating a Functional unit take the action of adding yoghurt as an example. This is illustrated as follows. First, place the yoghurt object in a bowl and the chopped banana in a mixer that is in turned off state. Then, perform the adding of the yoghurt action. The timing of the action is also mentioned in the motion state representation from the video. The object following the action is a mixer that has both chopped banana and yoghurt in it while the mixer is still in turned off state.

The syntax is as follows:

Object Line: O \t <Name of Object> \t <0 or 1 (describing if this object is moving or not)

State Line: S \t <Object's State> \t {List of ingredients, comma separated}

Motion Line: M \t <Name of Motion> \t <Starting time> \t <Ending time> (If the action is not there, then put Assumed)

The textual equivalent for Functional unit is as below:

```
//
O       Mixer    0
S       off
S       contains {chopped banana}
O       bowl     1
S       contains {yoghurt}
O       yoghurt  1
S       in       [bowl]
M       add yoghurt      1:46    1:49
O       Mixer    0
S       contains {chopped banana, yoghurt}
O       bowl     0
S       empty
//
```

This add yoghurt motion annotation is visualized as illustrated in Fig 1. The graph makes it evident that the motion has two inputs, a mixer in the off state with chopped banana and a bowl of yoghurt, and one output, a mixer in the off state with chopped banana and yoghurt.

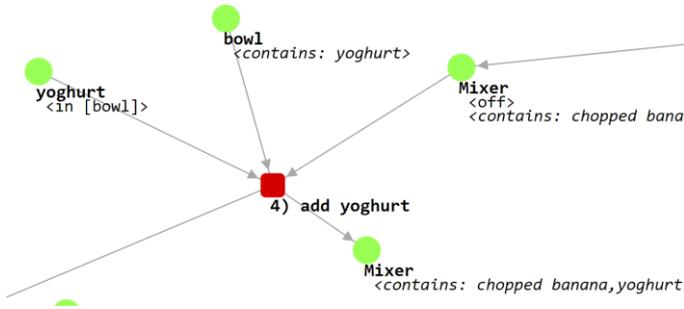

Fig 1: Functional unit of Add Yoghurt Motion.

From all the Functional units Universal FOON is created which can be used later for Knowledge Retrieval.

### III. METHODOLOGY

For creating FOON work plans for new recipes, the graph created from all the functional units can be used as the knowledge base. We can retrieve the task tree for the required task from the graph using particular search algorithms implement search methods to be able to extract a task tree for preparing any FOON dish. A task tree is structurally identical to a subgraph. Iterative Deepening Search and Greedy Best-First Search algorithms have been implemented. To extract the reference task tree, the Reference target Node must first be identified, and then the search techniques must be applied.

*A. Iterative Deepening Search Algorithm:*

A depth bound is established, and a depth-first search is carried out within of it. If the depth bound is 1, then DFS is carried out to depth 1 by treating all of the children of the start node as leaves. If a solution is not discovered, then DFS is carried out to depth 2. Up until the solution is found, keep performing DFS while increasing the depth. Every possibility must be explored in order to locate optimal solution because there may be several ways to prepare an object. You can just choose the first path you come across, though, to keep things simple. As searching for the answer, keep deepening your search. If the leaf nodes of a task tree are in the kitchen, it is regarded as a solution.

```
1  IDS(root, goal_node, depthLmt){
2      for d = 0 to depthLmt
3          if (DepthFirstSearch(root, goal_node, d))
4              return true
5      return false
6  }
7
8  DepthFirstSearch(root, d){
9      if root == goal_node
10         return true
11     if d == 0
12         return false
13     for child in root.children
14         if (DepthFirstSearch(child, goal_node, d - 1))
15             return true
16     return false
17 }
```

The space complexity is: O(bd), In this case, we suppose b is constant and that all children are formed at each depth of the tree and saved in a stack during DFS.

*B. Greedy Best First Search Algorithm:*

Heuristic function h(n) is used as a criterion to choose a path while investigating the node as opposed to randomly selecting a path from a variety of choices. It estimates the 'goodness' of node n, how close node n is to goal and the cost of minimum cost path from node n to goal state.
h(n) ≥ 0 for all nodes n
h(n) close to 0 means assume n is close to goal state
h(n) very big means assume n is far from goal state.

Two heuristic functions have been used to implement two Greedy BFS search strategies.

**Heuristics 1**: h(n) = success rate of the motion

If there are multiple paths with different motions, the success rates for each motion have been supplied in motion.txt file; select the one with the highest success rate to successfully execute the motion. For instance, say a robot has higher success rate of *adding* banana compared to *slicing* a whole banana i.e., adding motion has high success rate than slicing motion.

```
Input: Given Goal node G and ingredients I
T ← A list of functional units in Task tree.
Q ← A queue for items to search
Kingd ← List of items available in kitchen.
Q.push(G)
While Q is not empty do:
    N ← Q.dequeue()
    If N not in Kingd then:
        C ← Find all functional units that create
        C max ← -1 for each candidate in C do:
            if candidate.successRate > max then:
                max ← candidate.successRate
                CMax = candidate
        End for
        T.append(CMax) for each
        input in Cmax do:
            if n is not visted then:
                Q. enque(n)
                Make n visitied
            End if
        End for
    End if
End while
T.reverse()
Output: T
```

**Heuristics 2**: h(n) = number of input objects in the function unit

The path with least number of input objects is preferred for extracting the task tree. For instance say if cake mix can be prepared with either {egg, milk, sweet yoghurt} or {egg, milk, cheese, sugar powder, cardamom powder}, take the path that require {egg, milk, sweet yoghurt}. Because, in this path, you need fewer input objects.

## IV. EXPERIMENT/DISCUSSION

An iterative deepening search explores the FOON by performing DFS and BFS at the chosen depth bound. The depth level will continue to rise until a solution is found. If the answer emerges at a deeper level, this approach requires more time to assemble the task tree. This will add to the temporal complexity by traversing all previously visited nodes for each depth-bound increment. Because they follow BFS, heuristics 1 and 2 easily locate the answer at higher levels, but each complexity increases if the solution occurs at deeper layers.

TABLE I. NUMBER OF FUNCTIONAL UNITS FOR TARGET NODES

| Goal Nodes | Iterative Deepening Search | Heuristics 1 | Heuristic 2 |
|---|---|---|---|
| Sweet Potato | 3 | 3 | 3 |
| Ice | 1 | 1 | 1 |
| Whipped Cream | 10 | 10 | 15 |
| Macaroni | 7 | 7 | 8 |
| Greek Salad | 31 | 32 | 28 |

The task trees for all three methods could have the same or different numbers of functional units. All task trees have the same number of functional units for the target nodes ice and sweet potato.